Chapter 32

# Evolution in Virtual Worlds

Tim Taylor

Abstract

This chapter discusses the possibility of instilling a virtual world with mechanisms for evolution and natural selection in order to generate rich ecosystems of complex organisms in a process akin to biological evolution. Some previous work in the area is described, and successes and failures are discussed. The components of a more comprehensive framework for designing such worlds are mapped out, including the design of the individual organisms, the properties and dynamics of the environmental medium in which they are evolving, and the representational relationship between organism and environment. Some of the key issues discussed include how to allow organisms to evolve new structures and functions with few restrictions, and how to create an interconnectedness between organisms in order to generate drives for continuing evolutionary activity.

According to the neo-Darwinist theory of evolution, the richness and complexity of biological life can be explained in terms of three fundamental processes: reproduction, heritable variation, and competition for limited resources leading to natural selection. The beautiful simplicity of this picture raises the intriguing question: might it be possible to instill these processes in a virtual world, and, in so doing, unleash an ongoing evolutionary process that populates that world with a rich ecosystem of complex virtual organisms?

Attempts to do precisely this have a history as long as that of the modern digital computer itself. This chapter starts with a brief review of past work and the current state of the art; although much of this work is remarkable, the quest for open-ended evolution remains elusive; after an initial burst of activity, these systems tend to quickly reach a quasi-stable state beyond which no further qualitative changes are observed.

These results raise a nagging question: just how far can evolution progress in such worlds beyond what is easily discoverable by virtue of the specific way in which the world has been designed? The nature of these systems is examined in order to address this question. It turns out that such an analysis can tell us much, not just about evolution in virtual worlds, but also about the very nature of virtual worlds and the similarities and differences that exist between the virtual and the real.

In the latter part of the chapter, I pull together these ideas in order to map out the main components of a more comprehensive framework in which to study evolution in virtual worlds. This involves careful consideration of the desirable properties and representation of organisms and environment; a central issue here is how to design worlds in which the reproductive success of an organism depends upon its local environment, thereby promoting continual evolution. Considering the low-level design requirements to build a virtual world in which organisms and environment are richly interconnected could be described as a "bottom-up holistic" approach.

## Previous Work

In the late 1940s, von Neumann became interested in the question of how complicated machines could evolve from simpler ones (von Neumann 1966).[1] He wished to develop a formal theory of self-reproducing machines—machines that could build copies of themselves. Specifically, he was interested in self-reproducing machines that were robust in the sense that they could withstand some types of mutation and pass these mutations on to their offspring;

such machines could therefore participate in a process of evolution (Taylor 1999, 46–48). Looking for a suitable formalism that was both simple and enlightening, von Neumann developed a two-dimensional cellular automaton framework in which to demonstrate his ideas.

Although the design was not implemented on a computer before his untimely death in 1957, von Neumann's work can be regarded as the first attempt to instantiate an evolutionary process in the context of a modern, digital computational framework.[2] The work was seminal in setting out the logic of self-reproduction for evolving complex machines. A fundamental aspect of the design, which circumvented a potential infinite regress of description, was the dual use of information both to be interpreted as instructions for building a duplicate machine, and to be copied uninterpreted for use in the duplicate.[3]

Nevertheless, because of his focus on the logic of self-reproduction, von Neumann did not specifically deal with various other biological concerns, most notably regarding energy and the collection of raw materials.[4] Furthermore, he did not consider *interactions* between machines as a driving force for increased complexity. Rather, the little mention he did give to such interactions concerned their potential harmful effect in disrupting the functioning of self-reproduction within an individual machine. Von Neumann considered a system that had the *potential* for an evolutionary increase in complexity, but did not address the question of where the *drive* for such an increase may arise from within an evolutionary system itself.

However, some early implementations of computational evolutionary systems did consider interorganism interactions. Barricelli (1962; 1963), and Conrad and Pattee (1970) designed systems where mutualistic associations could arise between organisms. Although both systems exhibited some interesting ecological and evolutionary dynamics, attempts to evolve complex behaviors met with limited success. Conrad and Pattee remarked: "It is evident that the richness of possible interactions among organisms and the realism of the environment must be increased if the model is to be improved." They continued: "One point is clear, that the processes of variation and natural selection alone, even when embedded in the context of an ecosystem, are not necessarily sufficient to produce an evolution process" (407–408).

More recently, one of the most notable attempts to create a computational system in which natural selection leads to an open-ended evolutionary process has been Ray's *Tierra* (Ray 1991). This work studied the evolution of a population of self-reproducing computer programs, where the programs were written in a language based upon modern assembly code. The Tierran environment was a block of initially blank computer memory into which a single seed program, written by Ray, was placed. The program copied itself, one instruction at a time, into a new location in memory, and therefore created a new copy of itself; both copies then proceeded to reproduce, and so on until the memory filled up. When the memory was full, older programs were removed by the operating system to make room for new ones. Random mutations were sometimes introduced in the copying operations, such that variations emerged in the offspring programs. Ray observed that the programs evolved to reproduce more quickly, by optimizing their ancestral self-reproduction algorithm.[5] Furthermore, some of the most interesting results were due to ecological interactions; in particular, parasitism was seen to evolve, where short programs emerged that could only reproduce with the help of longer "host" programs. Resistance to parasites, "hyperparasites" (programs that subvert parasites for their own reproduction), and other related phenomena were also observed.

*Tierra* generated great interest within the nascent artificial life community in the early 1990s. However, as impressive as the results were, each particular run of the system would eventually reach a state of stasis in which only selectively neutral variations were seen to emerge (Ray 1992; 2011).

In 1993, inspired by *Tierra*, Ofria, Brown, and Adami developed a related system called *Avida*—for a recent overview, see Ofria, Bryson, and Wilke 2009. Unlike *Tierra*, where reproductive success ultimately boils down to how quickly a program can produce a copy of itself, programs in *Avida* can increase their rate of reproduction by performing specific, user-defined computational problems. *Avida* has been used to study the evolution of complex features (Lenski et al. 2003), but the drive for increased complexity was engineered into the environment by the authors via the provision of nine progressively more complex reward functions. Similarly, most of the other published studies with *Avida* have addressed specific topics either by making suitable adjustments to the reward functions (e.g., Elsberry et al. 2009) or by making targeted changes to the mechanisms for inter-program interaction (e.g., Beckmann and McKinley 2009). Thus, this work tends to be focused on evolving particular behaviors rather than addressing the question of how intrinsic drives for diversity and complexity can arise from within the system itself.

Taking a somewhat different approach, Holland developed a model called *Echo* that emphasizes the role of ecological interactions and exchange of resources in the evolution of complex adaptive systems (Holland 1995; Hraber, Jones, and Forrest 1997). *Echo* has been used for various studies involving ecological modeling (e.g., Schmitz and Booth 1996; Hraber and Milne 1997). However, its design is still restricted in terms of the evolvability of agents; the fact that the *Echo* operating system implicitly interprets the agents' behavioral specifications means that they can never come to encode anything more than the fixed range of actions (e.g., offense, defense, trade, and mating) predefined by the designer.

At around the same time as the original development of *Echo*, Yæger created a complex virtual ecology of evolving agents called *Polyworld* (Yæger 1994). In Yæger's system, agents controlled by genetically determined neural networks move around a two-dimensional environment, collecting energy, fighting, and mating. The agents are capable of a simple form of learning, and possess a relatively sophisticated vision system where visual input is determined by a rendering of the scene from an individual agent's point of view. In addition, physical obstacles and barriers can be placed in the environment to restrict the agents' movements. Yæger presented a qualitative description of results, in which it appeared that distinct species of organisms evolved and coexisted. However, evolvability is still restricted by the fact that interagent interactions are drawn from a small set of primitive behaviors (move, turn, eat, mate, attack, light, and focus).

Perhaps the most visually impressive work on evolution in virtual worlds to have been conducted to date has been that of Sims, together with more recent related work by other authors (Sims 1994b; 1994a; Ventrella 1999; Taylor and Massey 2001; Lassabe, Luga, and Duthen 2007; de Margerie et al. 2007; Miconi 2008). Sims allowed the body shape and movements of three-dimensional creatures to evolve at the same time, in a virtual world featuring simulated Newtonian mechanics. Each creature is built up from a genetic description that describes both its morphology and its control architecture. This representation provides modularity to the mapping from genotype to phenotype, and naturally leads to features such as duplication and recursion of body parts. In some runs, the creatures lived in a simulated fluid medium, and, in others, they lived in a terrestrial environment with gravity and a ground plane. In contrast to most of the previously discussed work, Sims used a traditional genetic algorithm with fitness functions designed to reward specific behaviors (such as moving forward, or following a target) rather than employing self-reproduction and open-ended evolution. Some example creatures evolved by Taylor and Massey (2001), inspired by Sims's original system, are shown in figure 32.1.

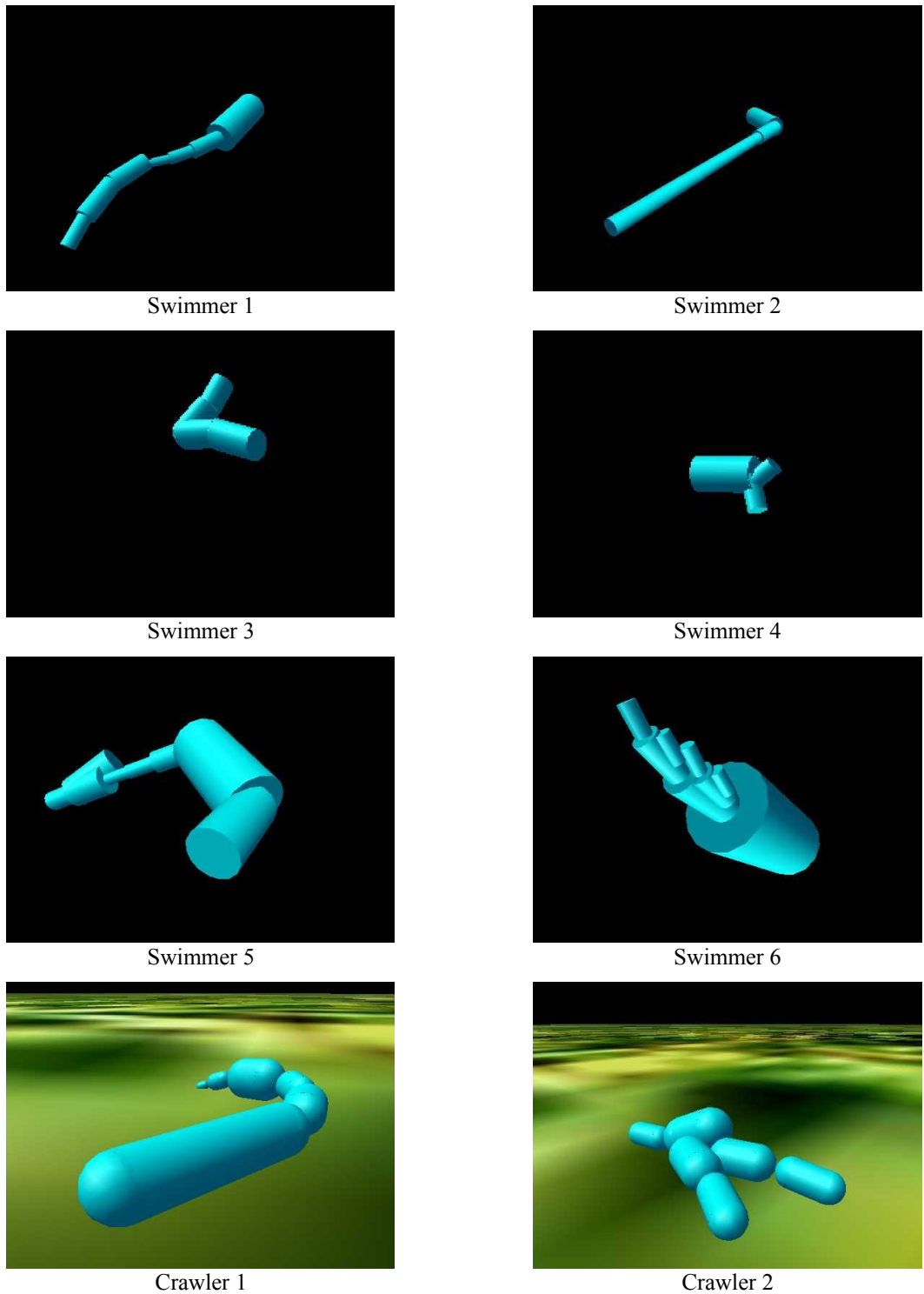
Swimmer 1  Swimmer 2
Swimmer 3  Swimmer 4
Swimmer 5  Swimmer 6
Crawler 1  Crawler 2

**Figure 32.1:** Some virtual creatures evolved by Taylor and Massey (2001).

One of the reasons that Sims's system produced such good results was that he modeled the physics of a three-dimensional environment accurately enough that objects moved realistically when subjected to forces and torques. Hence the beautiful movements produced by many of his evolved creatures were due just as much to the accurately modeled physical environments as they were to the creatures' individual controllers. In some of his later work, Sims (1994a) looked at evolving pairs of opponents to compete in simple games (involving fighting for possession of a free moving cube); work that graphically demonstrated how coevolutionary arms races (Dawkins and Krebs 1979) can lead to complex morphology and behavior.

Several other authors have attempted to move away from explicitly defined fitness functions to create virtual worlds with simulated physics where creatures may evolve in a more open-ended fashion. Earlier work was performed in two-dimensional worlds (e.g., Ventrella 1999) and more recent work in 3D (e.g., Miconi 2008). However, the computational complexity of the simulations only allowed for populations of a couple of hundred creatures, and the evolutionary results reported so far have been fairly restricted.

## Open Problems

It is clear from the preceding review that work on evolution in virtual worlds has not yet succeeded in reproducing the long-term evolutionary dynamics observed in the biological world. Although much of this work is remarkable, none has achieved an open-ended evolutionary dynamic involving a long-term, intrinsic drive for increased diversity and complexity of the virtual organisms. One conceivable explanation is that the scale of these systems, both in terms of population sizes and durations of runs, has simply not been large enough to date; if a much larger system were run for a much longer time, perhaps we would see more interesting evolutionary phenomena emerge. However, there are a number of reasons to believe that the poor evolvability is due not just to issues of scale, but also to some more fundamental problems with the way in which these systems have been designed. Some of the most apparent of these issues are highlighted below. Consideration of the results of work to date, in the light of such issues, suggests that the processes of self-reproduction with heritable mutation and selection, by themselves, are insufficient to explain the open-ended evolution of diversity and complexity.

### Fitness

In much of the work described above, there was a conscious attempt to avoid defining an explicit rule—a "fitness function"—to determine which individuals were allowed to reproduce. It has often been argued that avoiding an explicit fitness function is a key ingredient for achieving open-ended evolution (e.g., Packard 1988; Miconi 2008). A common way to accomplish this has been through self-reproduction—requiring organisms to build their own offspring rather than employing an extrinsic mechanism to decide which organisms can reproduce. Describing the design of *Tierra*, Ray explained:

> [Self-reproduction] is critical to synthetic life because without it, the mechanisms of selection must also be predetermined by the simulator. Such artificial selection can never be as creative as natural selection. The organisms are not free to invent their own fitness functions. Freely evolving creatures will discover means of mutual exploitation and associated implicit fitness functions that we would never think of. Simulations constrained to evolve with predefined genes, alleles, and fitness functions are dead-ended, not alive. (Ray 1991, 372)

However, the situation is somewhat more complicated, because in order to "discover means of mutual exploitation," the system must allow the evolution of new forms of *interaction*, and the requirement of self-reproduction by itself is not sufficient to ensure this. The question of evolving new forms of interaction is discussed in the following section. Furthermore, some authors have argued that even in virtual worlds with self-reproducing organisms, there will always be some aspects of the reproduction process that have to be designed a priori by the programmer (e.g., Miconi 2008). However, I argue later in the chapter that the degree to which this is true depends on how the distinction between organisms and environment is represented in the virtual world.

## Restricted Ecological Interactions

The most interesting evolutionary innovations to emerge in *Tierra* were those that involved interactions between different programs, such as parasitism, immunity to parasites, hyperparasites, and so on. However, the range of interactions that could emerge was restricted to those that were possible given the specific "interaction enabling" features of the language in which the programs were written; these allowed a program to search for a particular location in a neighboring program, and to read or execute code from that location. These facilities enabled certain types of interaction (mostly related to parasitism and related phenomena), but did not allow for the appearance of many other conceivable interactions.

Interorganism interactions in most of the other work discussed above were even more restricted. An interesting exception was the work of Sims on evolving pairs of opponents to compete in games in a three-dimensional virtual world. Here the interactions between the opponents were mediated through the creatures' bodies, modeled as physical structures in an environment with simulated Newtonian mechanics. These environmental dynamics afforded the potential for a rich variety of possible interactions. The resulting coevolutionary arms races produced some of the most impressive results for virtual evolution yet observed.

Another aspect of opening up the potential range of interactions between organisms is allowing for the evolution of new sensors and effectors.[6] These provide the two directions of influence between environment and organism across the organism's boundary, and the evolution of these capacities is difficult in a computational medium because the representation of this boundary is usually hard-coded and immutable. However, without such evolution, these systems are confined to evolving complex computational processing on the sensory information provided by the system designer—they are unable to evolve new forms of input and output in order to exploit other properties of the environment. This topic will be returned to below.

In some of the other work previously discussed, such as *Echo* and *Polyworld*, the environments contained material resources that organisms had to find and collect in order to survive and reproduce. This introduced the possibility of indirect interactions between organisms, where the availability of resources in the environment for one organism could be affected by the behavior of other organisms. The evolutionary potential of these systems still suffered from the organisms having a fixed set of actions available to them. In the biological world, organisms have to collect the materials and energy required to create their offspring, as well as to maintain their own structure. This direct link between uptake, transformation, storage, and excretion of resources, on the one hand, and survival and reproduction on the other hand, is missing in all of the systems described above, and I return to these issues below.

## Fixed Representation and Structure

An issue common to the majority of systems discussed above is that the basic structure of an organism is fixed. For example, a Tierran organism always consists of a string of code (the program) together with various elements that together define the state of its "virtual CPU" (namely, four registers, a stack, and an instruction pointer); in *Echo*, an organism consists of a chromosome that defines its behavior, and a reservoir in which it stores any resources it has acquired from the environment. In the biosphere, the most dramatic moments in evolutionary history have been the so-called major transitions (Maynard Smith and Szathmáry 1995), in which the very structure of an organism has radically changed (e.g., the transition from unicellular to multicellular life). Such changes are not possible in virtual worlds in which the scheme for representing an organism is not itself mutable.

A related issue is the very direct, and fixed, relationship between "genotype" (an organism's hereditary material) and "phenotype" (an organism's physical presence and behavior in its environment) in some of these systems. This issue arises when the machinery that processes the genotype (e.g., the virtual CPU in *Tierra*) is not itself evolvable. Without the possibility of evolving new ways to decode the genotype into a working phenotype, there is no chance of evolving different, and potentially better, ways of representing complex phenotypes.

## Lack of Complex Dynamics in Environment

One of the key aspects missing from all of the previously described work, with the exception of that of Sims and related studies, is an environment possessing its own complex dynamics. In most of these systems, the environment is essentially an inert medium that provides a space in which organisms can exist, in some cases with resources and other items. As already noted, the lifelike movements displayed by Sims's evolved creatures were a result of the interaction of the creatures' limb movements and the simulated Newtonian dynamics of the environment.

An environment can potentially provide many different functions, such as force fields that determine how objects move, various mechanisms for the transmission of information, determining how objects interact, and so on. To date, very little attention has been given to how the properties of the environment affect the evolution of complex organisms. These issues will be further discussed below.

Furthermore, it is widely accepted that at least some of the mass extinction events in the history of biological life were caused by external shocks such as meteor impacts (Raup 1986), and yet few virtual worlds model such catastrophes induced by the abiotic environment. However, it has been argued that most extinction events, and the continual turnover of species that result from them, may be caused by the intrinsic dynamics of the evolutionary process itself (Solé et al. 1997). Whether or not external shocks are required to promote continued large-scale evolutionary change remains an open question.

## Restricted Population Size and Structure

Most of the work reported above could cope with population sizes of a few hundred individuals, or a few thousand at most; *Avida* is capable of running the largest populations, up to around 12,000 individuals in recent work (Elsberry et al. 2009). It is likely that the evolutionary potential of these systems is significantly restricted because of these small population sizes. In the biological literature, the concept of "minimal viable population" (MVP) refers to the lower bound on population size such that a species can survive in the wild. Recent surveys suggest a median MVP value of approximately 5,000 individuals (Traill, Bradshaw, and Brook 2007).[7]

Theoretical reasons for a minimum viable population size include inbreeding and lack of genetic diversity, and demographic and environmental stochasticity. Furthermore, if a system is to accommodate food chains of species at different trophic levels, many individuals of the species at the lower levels are required to provide sufficient food for species at higher levels. While it can be dangerous to apply empirical results from the biological world directly to virtual worlds, these factors do serve as a warning that the limited capacity of many virtual worlds to support large population sizes may be a problem.

It should also be noted that, in much of the existing work, organisms reproduce asexually—there is no mixing of genetic material between individuals either "vertically" (through sexual reproduction) or "horizontally" (the exchange of genetic material between unrelated organisms). Although some attempts have been made to introduce sexually reproducing organisms into these worlds (e.g., Taylor 1999), evolution of such populations tends to result in the emergence of simpler, asexually reproducing variants that eventually replace the sexually reproducing individuals. Both vertical and horizontal gene transfer are common in biological life and have significant, if not fully understood, consequences (e.g., Hurst and Peck 1996; Doolittle 2000). The omission of such processes in current work on virtual evolution is therefore likely to be a substantial source of divergence from the dynamics of biological evolution.

# Components of a More Comprehensive Framework

One reason for the limited results of past work is often an overemphasis on the requirements for a Darwinian evolution process to the exclusion of other aspects of biological theory. In particular, much of the work pays very little attention to ecological processes such as food webs and resource cycles. As will be discussed later, it is likely that such processes play an important role in promoting the open-ended evolution of diversity and complexity.

However, it is also apparent from the analysis above that there are other important issues to be addressed, beyond those traditionally tackled in the fields of theoretical biology and ecology. These include the design of the environment and the representational relationship between organisms and environment. Such questions seldom arise in traditional biological theory because the nature of the physical and chemical world can be taken as a given. However, when designing virtual worlds, we must explicitly design all aspects of the world; careful thought must go into this design if we wish to produce a world in which an open-ended evolutionary process may unfold. Here, I pull together these ideas in order to map out the main components of a more comprehensive theoretical virtual biology.

## Design Goals

It should be emphasized that the following sections describe many different aspects of the design of a virtual world that might support an open-ended evolutionary process. A substantial research effort is required to make progress in these areas. In reality, at least in the near future, the design goals of specific virtual worlds are likely to be more narrowly defined; hence, some aspects of the following will be more immediately relevant than others.

Some examples of possible objects of study include the following:

- A focus on the origin of living systems and the evolution of basic autonomy versus a focus on agents with "higher-level" intelligent behavior involving processes such as learning, memory, communication, and language
- Evolution in "native" digital environments with discrete memory locations and discrete execution of instructions (e.g., Internet agents) versus evolution in simulated physical environments with (simulated) continuous time and space
- Guided evolution to produce agents for specific purposes versus open-ended evolution of diverse, complex organisms

In the following sections, I discuss the relevance of each topic in relation to each of these goals. Much of this concerns the design of virtual worlds that can support open-ended evolutionary processes with as few restrictions as possible as to what can evolve. This necessarily requires us to focus for the most part on basic, low-level design features. If the design goal of the system is to evolve organisms with higher-level, more human-like intelligence, then it may make sense to forgo some of the complete freedom in evolvability of organism structure, and concentrate on specific mechanisms designed to aid the evolution of features such as learning, memory, and communication. However, further discussion of such issues is beyond the scope of this chapter.

## Nature of the Individual

What constitutes an appropriate representation for an individual organism will depend upon the design goals of the virtual world. In the biological world, organisms are continually engaged in the procurement of matter and energy, not just to reproduce, but also simply to survive and maintain their own structure. Thus, organisms are the connecting tissue of twin hierarchies—an evolutionary hierarchy (involving levels such as genes, organisms, and species) and an ecological hierarchy (involving levels such as organisms, ecosystems, and the global biosphere) (Eldredge 2008).

In the context of the design of a virtual world, the notion of an organism as an ecological actor presents a variety of issues, particularly the modeling of food chains (and associated processes of capture, storage, transformation, use, and exchange of resources), and the representation of the organism's structure.

The concepts of food chains and webs, as used in the biological literature, only make sense in virtual worlds in which organisms are composed of atomic elements that are subject to a law of conservation. In von Neumann's cellular automata model, for example, and in *Tierra* and *Avida*, organisms could create copies of themselves "out of thin air," without having to collect the individual components required to build the copy from elsewhere in the environment. Hence, in these worlds, there is no requirement for, or possibility of, the emergence of food chains. The consequences of this will be discussed in the following section.

Some of the other systems discussed previously, such as the work of Conrad and Pattee and of Holland, did require the organisms to collect resources in order to reproduce.

However, in these systems the requirement to collect resources was not directly connected to the composition of the organism itself but was essentially arbitrary. This arbitrariness arises because the organisms are not fully embodied in their virtual worlds—their representation is distinct from that of the environment. I discuss later the consequences of this lack of embodiment, in terms of the evolution of ecosystems and of the evolution of an organism's own structure.

Whether or not the organisms are fully embodied in the virtual world, the nature of their genetic information—the inherited information passed from parent to offspring—must be carefully considered. Von Neumann's work on self-reproducing automata addressed the issue of how to ensure the availability of pathways in the space of possible genomes to allow evolution to move from simple to complicated organisms. His proposed architecture, upon which his self-reproducing automata are based, is a solution to the problem, and gives the automata the potential to evolve into progressively more complicated forms. However, the design of systems such as *Tierra*—in which programs reproduce simply by copying themselves one instruction at a time, with no strict genotype-phenotype distinction—suggests that von Neumann's full architecture is not always required for the evolution of complexity. In the case of *Tierra*, programs can reproduce in this manner because they are one-dimensional structures where each element can be easily accessed in order to be read and copied. In the two-dimensional environments considered by von Neumann, such a strategy would not be possible in general. We can therefore say that the self-reproduction architecture required in order to allow for the evolution of complex organisms will depend on the nature of the medium—in particular, on its dimensionality and dynamics. More work is required to fully understand these dependencies.

In addition, the mechanisms for replication and mixing of genetic information, both vertically and horizontally, must also be considered. Ideally, it should be possible for new mechanisms for genetic mixing to evolve, and this again points to the desirability of allowing an organism's structure to be subject to evolution; I deal with this point below.

Finally, any virtual evolutionary system must be seeded with some designed structure —an ancestral organism—to start the evolutionary process. The choice of a suitable seed structure will depend upon the design goals of the system. To recreate the origin and early evolution of life, imposing few assumptions on what might emerge, an appropriate seed might be a simple self-replicating structure with the ability to initiate other dynamics in the world.[8]

If, however, the focus of the system is on the evolution of higher-level intelligence, then it may be desirable to start with a more complex ancestor that already has some assumptions and capacities for information processing, communication, and learning.

## Nature of the Ecosystem

The general lack of support for complex ecosystems in existing virtual worlds has already been highlighted. Organisms in systems such as *Tierra* and *Avida* compete for CPU time to execute their instructions (which the authors of these systems regard as a metaphor for competition for energy), and they also compete for limited space in memory in which to build their offspring. However, as already noted, the matter from which they are composed can be created out of thin air (it is not conserved), and is therefore not something for which organisms compete.

There are several consequences that arise from this lack of competition for building blocks. First, in addition to a lack of competition between organisms for resources in the environment, organisms are not themselves resources of matter for other organisms; a program in *Tierra* can read an instruction from a neighboring program, but it does not need to (and indeed is unable to) actually *remove* instructions from the neighbor in order to build its offspring. Although a program can read and execute useful code from a neighboring program (we might say that the neighbor is acting as a resource of *information*),[9] there is no life-or-death struggle between organisms over the very building blocks from which they are composed. Hence the coevolutionary pressures on the organisms to develop increasingly elaborate defenses and weapons are much weakened, if not totally absent.

Second, in the biosphere, the conservation of matter, and the resultant cycle of resources that this necessitates throughout an ecosystem, creates an underlying *interconnectedness* between all members of the ecosystem. Organisms are consumers and producers of resources, and the existence of one species creates opportunities for other species to exist (e.g., ones that feed on it, or which decompose its waste). Furthermore, the interconnectedness of ecosystems means that the loss of one species may have significant ecological and evolutionary consequences for many other species in the system. Hence, the lack of competition for material resources in virtual evolution systems is probably a significant contributory factor to their lack of continued evolutionary activity and their low diversity of species.

In addition to considering material resources, the role of energy, or its equivalent in virtual worlds, must also be considered. Above, it was suggested that CPU time in *Tierra* and *Avida* might be regarded as an analogy to energy in biological systems. But energy in the physical world is, of course, a much richer concept; at the chemical level, it determines which chemical reactions can happen and when and, at the physical level, it allows organisms to deploy stored energy as useful work, acting against an external physical force and thereby exhibiting a degree of autonomy. Whether or not it is appropriate to model such properties in a virtual world will depend on the design goals for the system.

When designing a virtual world, decisions must be taken about how to model energy and material resources, and the rules that govern the reaction, transformation, capture, storage, and transmission of materials. These decisions will depend on whether one is trying to simulate physical systems or to work in a more native computational domain (or somewhere in between these two extremes). As explained above, the decisions taken will have significant consequences for the evolutionary behavior of the system—although the precise nature of these consequences remains to be elucidated. It is therefore important that the decisions are carefully considered and related to explicit motivations derived from the design goals, rather than being treated as a mere implementation detail.

## Nature of the Medium

Perhaps more than any other aspect, the nature of the medium in which the evolving virtual organisms live has received very little explicit discussion in previous work. The medium is the shared area in which organisms and abiotic objects act and interact. It defines the concepts of space and neighborhood. In addition, it defines any global dynamic processes that act on all objects contained within it (the "laws of physics"), and hence also defines a global concept of time. As I discuss in this section, many of the virtual evolutionary systems we have considered also have predefined areas of space specifically associated with individual organisms; these do not exist in the shared medium and are therefore not subject to the global laws of physics. Similarly, many systems also have local update procedures specifically associated with individual organisms rather than applying to all (biotic and abiotic) objects in the medium. Indeed, some systems *only* support these local update procedures for organisms, which therefore exist in an inert medium possessing no global laws of physics.

The nature of the medium is generally not discussed in traditional theoretical biology, as the properties of the physical world can be taken for granted. But the evolutionary phenomena that might be expected to arise in a system are intimately related to the properties of the medium in which the evolutionary process is unfolding, as will be highlighted in this section. Hence, it is vital that these properties are carefully considered when designing a virtual world.

## Discrete and Continuous Media

In some of the preceding discussion, a distinction has been drawn between "native computational" environments (such as those provided in *Tierra* and *Avida*), and simulated physical environments (such as those provided in the work of Sims and related studies). One component of this distinction is whether the space in which organisms live is discrete or continuous. In practice, assuming the world is implemented on a digital computer, the space must be discrete at some level, as the position of an object cannot be specified to an infinitely fine level of detail. Thus, in practice, this component is in fact a continuum of "granularity of discreteness" rather than a discrete-continuous dichotomy. The same comments also apply to the representation of time in the virtual world.

## Embodiment and Evolvability

A more relevant distinction in the current context is the algorithm by which the state of the world is updated. This may operate at the level of the smallest elements of the world (e.g., an update rule for an individual cell in a cellular automaton) or it might operate on higher-level constructs. For example, in *Tierra*, the state of the world is updated by the "virtual CPU" possessed by each live program. Each program's virtual CPU decides which instruction to execute at the current time step. In Sims's virtual creatures, there is a multistage update algorithm, in which a creature's controller is first updated to determine the forces to be applied by each of its joint actuators at that moment; then the simulation of Newtonian mechanics is updated to determine the resultant movement of the creature.

The important point is that, in any virtual world in which the update algorithm operates on anything other than the smallest elements of the world, a design decision has to be made about which higher-level constructs to act upon. This then "hardwires" the notion of these higher-level constructs into the design of the system itself. In work on evolution in virtual worlds, these higher-level constructs are, of course, usually the organisms themselves. If the state of the world is being updated at the level of the organism rather than lower-level elements, the system must be able to identify and keep track of the organisms. This necessarily requires a predefined representational distinction between organism and environment and means that some aspects of the organism's structure are not embodied in the medium.

For example, in *Tierra*, an organism is defined as a string of instructions together with the various elements associated with its virtual CPU (i.e., its registers, stack, and instruction pointer). But only the string of instructions is embodied in the shared medium of the world—the Tierran memory space—and potentially accessible to other organisms. Furthermore, although Tierran organisms have to copy their instructions into a new spot in the environment in order to reproduce, they do not have to copy their registers, stack, and so on; these items are automatically replicated by the system when an organism reproduces rather than having to be explicitly copied by the organism itself.

Similarly, in Sims's work, only a creature's limbs exist in the environment as simulated physical bodies. Its controller, actuators, sensors, and genetic description and decoding mechanism are not represented as physical entities in the environment. Instead, they are composed of predefined components that are not themselves evolvable. As a consequence, a creature could never evolve a new method of producing itself from its genetic description (a new genotype-phenotype mapping), nor could it evolve new types of sensors or actuators.

Such a predefined representational distinction between organism and environment therefore introduces serious consequences for the evolvability of the system. Because the basic design of an organism has been predefined, it is not itself able to evolve; a program in *Tierra* could not experience a major evolutionary transition in its architecture to become a multiprocess parallel program—unless such a capacity was explicitly programmed into the system by the designer, as was the case in Thearling and Ray (1994). And yet, as mentioned earlier, these kinds of major transitions in the organization of individual organisms have marked key moments in the evolution of complex biological life (Maynard Smith and Szathmáry 1995).

Certainly, those components of an organism that are not represented within the shared medium (such as a Tierran organism's registers or a creature's actuators in Sims's system) *could* evolve if the system was so designed. The point, however, is that these components are not constructed by the organism itself when it is building its offspring; the mechanism for their reproduction, and the potential ways in which they could evolve, must therefore be predefined by the designer. Hence, such components could still only evolve in certain predefined ways.

Furthermore, a predefined representational distinction between organism and environment implies the existence of a boundary between the two to demarcate what does, and does not, belong to an organism. If the organism is to do anything in the world, this further entails predefined mechanisms for specific cross-boundary processes, possibly involving the transport of resources or the transmission of forces or information. But, again, if these must be predefined, then the ability to evolve new cross-boundary processes (e.g., new sensors or effectors) will be absent or, at best, only evolvable in certain predefined ways. I will return to this topic shortly.

## Interconnectedness through the Properties of the Medium

Returning to the nature of the algorithm that updates the state of the world, there are other aspects of its implementation that also have important consequences for the evolutionary potential of the system. In a "computational-like" medium like *Tierra*, the elements are discrete memory locations containing state information that is treated as instructions or data or both. Memory locations are inert unless specifically acted upon by an instruction; the modes of interaction in such systems, mediated by specific instructions, therefore have to be explicitly designed into the system.

In contrast, in worlds with simulated physics, the medium supports dynamics that act upon all elements, such as gravity and fluid drag forces (in Sims's work), and the transmission of visual information (in *Polyworld*). Hence, objects in simulated physical worlds are continuously affected by the presence of other objects in the world, without having to actively initiate interactions. They are bathed in a sea of information providing a potentially rich Umwelt and representing another form of interconnectedness between organisms in addition to that provided by the existence of an ecosystem of resources.[10] Such dynamics provide rich possibilities for interorganism interactions, as discussed above. In contrast, objects in a computational medium are blind to their surroundings unless they utilize specific mechanisms for communication that have been predefined by the designer.

## Evolution of New Sensors and Effectors

Simulated physical worlds may support phenomena in one or multiple domains; the domains of Newtonian dynamics and transmission of light have already been mentioned, but any number of other domains of physical phenomena could also be implemented, in addition to phenomena that have no analogues in the real world. In virtual worlds with simulated physics, the medium of the environment therefore inherently exhibits complex phenomena, and an important aspect of the evolution of complexity concerns the question of how organisms can evolve to capture and exploit these phenomena for their own benefit. The designer of a virtual world must provide the organisms with some tools with which they can sense and influence their surroundings—that is, with some sensors and effectors. In Sims's work, for example, a fixed set of different actuators is available for use in an organism's joints. Each sensor or actuator will work with a particular domain of phenomena (e.g., the joint actuators work in the domain of Newtonian dynamics, and light sensors work in a very simplified version of the domain of electromagnetic radiation).

An important aspect of open-ended evolution is how organisms can evolve to do things beyond what has been "programmed in" to the system by the designer. This relates not just to evolving complex information processing tasks, but also to evolving new ways of interacting with the world—new sensors and effectors. Within a single domain, new forms of action might arise if an organism evolves to initiate progressively more complex chain reactions of dynamics in the environment. However, if the environment has multiple domains of phenomena, we face the additional problem of how organisms might evolve to capture phenomena in a new domain in which no sensors or effectors have been predefined by the designer. Ultimately, this must come down to (at least some) components in the system having multiple properties across different domains, which can act as bridges from one domain to another. In an evolutionary context, an organism might have evolved to make use of a component because of its properties in one domain (e.g., its ability to act as an actuator), but other properties of the same component may subsequently become useful and be selected for (e.g., the same component may also be sensitive to light, and therefore act as a rudimentary eye). Hence, in worlds in which components exist that can act as bridging technologies across multiple domains of phenomena, organisms can evolve new forms of sensors and effectors beyond those programmed into the system by the designer.

## Physical Ecosystem Engineering and Niche Construction

The capture and use of food and energy by organisms was considered in a previous section. However, biological organisms utilize many aspects of their environment beyond those that provide food and energy. Nontrophic resources may be useful to an organism in a multitude of ways, by making its life easier or less dangerous in some way. Examples include resources that help regulate the environment (e.g., providing shelter or protection); tools to help with the capture, preprocessing, storage, and transport of other resources; tools for offense and defense against other organisms; tools to extend an organism's capabilities for signaling and communication, and so on.

By using resources in the environment in this way (a process known in the literature as "physical ecosystem engineering" or "niche construction"),[11] an organism's behavior can have significant ecological and evolutionary consequences for other organisms of the same or different species (Jones, Lawton, and Shachak 1997; Odling-Smee, Laland, and Feldman 2003). For example, species that build nests for their offspring reduce the selection pressure on the offspring's ability to withstand harsh environments by buffering environmental variation. Another example is provided by the dam-building activity of beavers that drastically alters the local environment experienced by the beavers and many other species in a way that can last for many generations (Naiman, Johnston, and Kelley 1988). In this situation, there is a "reciprocal causation" in the relationship between organism and environment; changes to the environment caused by the action of a species can alter the selective environment acting upon the same or other species and therefore affect how they evolve (introducing a form of "ecological inheritance" in addition to genetic inheritance).

If the medium of a virtual world is endowed with nontrophic resources that can help organisms survive in some way, similar processes of physical ecosystem engineering and niche construction can be expected to emerge. These processes provide another level of interconnectedness between organisms in the environment, such that changes in the behavior of one organism will affect other organisms and thereby potentially promote continued evolutionary activity. In addition, heterogeneity in the environment, which could result from processes such as niche construction, can lead to spatial segregation of organisms. In time, this can lead to isolated populations, thus promoting speciation and diversity within the system.

# Bringing It All Together: Embodiment, Self-Reproduction, Interconnectedness, and Open-Ended Evolution

The considerations in the previous section help elucidate the relationship between the concepts of fitness, self-reproduction, and open-ended evolution. If some parts of the organism are reproduced automatically according to a predefined mechanism (i.e., not embodied in the medium), there must be a predefined procedure to decide *when and how* such a mechanism operates. Such parts will therefore not be subject to variation and evolution or, at best, only subject to evolve in certain predefined ways. That is, in order to avoid any hardwired restrictions on evolvability, the organisms must be *fully embodied* in the shared medium of the world. Full embodiment entails an organism being composed solely of components that are subject to the general laws of physics of the medium and are not subject to any special higher-level update rules. Full embodiment therefore *necessitates self-reproduction*, as it entails that there are no special ancillary processes to aid in the identification and reproduction of organisms. Of course, depending on the design goals of the system, one might forgo total evolvability to more easily achieve particular outcomes.

The concept of a fitness function can be viewed as the determination of whether, and when, an organism can reach a state where it can reproduce. Hence, for a community of organisms, it defines a driving force that influences the current state and direction of change of the composition of the community. For open-ended evolution, we wish to avoid fitness functions that define static fitness landscapes, as these imply optimal states beyond which no further evolution is possible. The way to avoid static fitness functions is to make the fitness of an organism dependent not just on the organism itself but also on its local environment (which may include other organisms). This can be achieved if the medium creates interconnectedness between organisms, through the creation of food webs, through dynamic processes supported by the medium such as the transmission of forces or information, or through niche construction. Such a dependency will introduce coevolutionary drives and dynamic, shifting fitness landscapes.

To summarize, the degree of embodiment of an organism in the medium dictates which aspects of the organism are evolvable rather than hard-coded. By definition, those parts that are embodied must be constructed by the organism itself when it is building its offspring. Hence, there is a close relationship between embodiment and self-reproduction, and the degree to which these are present determines the extent to which an organism can freely evolve without predefined constraints. For self-reproducing organisms, the variety of possible forms is also clearly determined by the properties of the medium and its capacity to support complex arrangements of components and dynamic processes of action and interaction between components. These aspects define the set of potential organisms, but even with a large set of possibilities, the ability of the evolutionary process to traverse the genetic space from one to another may still be restricted. This is precisely the problem that von Neumann tackled, and for which his genetic architecture is a solution (but perhaps not the only solution). Having considered the diversity and connectivity of the space of potential organisms, the drive for evolution must also be considered. This comes from the decision on which organisms can reproduce and when (i.e., the concept of fitness). If this depends solely on the organism itself, it will lead to a static fitness landscape and the likelihood of eventual stasis in the population. If, however, fitness depends on the organism and its local environment, a dynamic fitness landscape will arise, with opportunities for continual evolution. This can come about through the interconnectedness between organisms provided by food webs involving abiotic or biotic resources, by dynamic processes of interaction and communication supported by the shared medium, or by physical ecosystem engineering and niche construction.

The organisms in *Tierra* are self-reproducing, but they are not fully embodied, so the structure of organisms that can evolve is restricted. A limited, unidirectional connectedness is allowed by organisms being able to read (but not write) the code of neighboring organisms. Of the systems discussed earlier, only those of von Neumann and Barricelli are fully embodied. However, neither of these worlds support laws of conservation of matter, and hence they lack the notion of food webs and the associated interconnectedness between organisms and coevolutionary dynamics that arise from them. Although Barricelli observed many interesting results, his virtual world is also hampered by the fact that the evolutionary process unfolds in an inert computational medium.[12]

The organisms in both Barricelli's and von Neumann's systems turned out to be very sensitive to perturbations from the environment. This is a particular problem with von Neumann's organisms, which are vastly more complicated than those studied by Barricelli. This raises the caveat that if an organism is fully embodied in the shared medium of the world, it must *engage in maintaining its own structure* so that it can survive perturbations from the environment for long enough to enable it to reproduce. Thus, a major challenge in future work is to create a system in which fully embodied organisms actively maintain their own structure[13] while still fulfilling the other requirements for open-ended evolution discussed throughout this chapter.

As demonstrated in the preceding discussions, the pursuit of open-ended evolution in virtual worlds requires synthesizing knowledge not just from a narrowly defined view of neo-Darwinism, but also from the wider literature on theoretical biology, in addition to addressing more technical concerns. By making advances in the various areas outlined here, in the near future we can expect to see significant improvements in the evolutionary potential of virtual worlds to produce diverse ecosystems of complex virtual organisms.

# Notes

1. This chapter focuses on the technical challenges of instantiating evolution in virtual worlds. For cultural and philosophical perspectives on the history of artificial life, see Riskin 2007; Johnston 2008. Discussion is omitted of popular computer games such as *Spore* (Electronic Arts, 2008), as these generally model evolution at a very superficial level (Bohannon 2008).
2. A working implementation, based upon von Neumann's design with some minor changes, was developed more recently by Umberto Pesavento and Renato Nobili (Pesavento 1995).
3. Von Neumann's description of the logical design of a self-reproducing machine can equally be applied to the reproductive apparatus of biological cells. However, although his work predated the unraveling of the details of DNA replication by some years, it had little impact on developments in genetics and molecular biology (Brenner 2001, 32–36).
4. Von Neumann had originally intended to return to these issues later on (von Neumann 1966, 83, 93–99).
5. The optimizations came about by the natural selection of variant programs, introduced by random mutations, which required less CPU time to effect their replication. This could be achieved by finding ways to reproduce with fewer instructions (as fewer instructions to copy meant a faster replication rate); Ray observed the evolution of self-replicating programs that were barely one-third of the length of his original handwritten ancestor. Alternatively, in other runs he observed programs that had evolved more sophisticated copying algorithms that could copy a given size of program using fewer CPU cycles than the original ancestor (Ray 1994).
6. The terms *effector* and *actuator* are both used in this chapter, and have slightly different meanings. An effector is a device that causes a change in the environment (e.g., a wing can cause flight when suitably controlled). An actuator is a device that actually provides motive power (e.g., a muscle). An effector will therefore contain at least one actuator as a subcomponent.
7. Although see Garnett and Zander 2011; Brook et al. 2011 for further debate on this topic.
8. For a full discussion, see Taylor 1999, §7.2; and Taylor 2001.
9. This feature was exploited by the evolved parasites discussed earlier.
10. Hoffmeyer (2007) provides an interesting elaboration of these issues from the perspective of biosemiotics.
11. The term "niche construction" actually refers to a broader category of phenomena whereby organisms modify the environment that they experience. This includes changes to trophic, as well as nontrophic, aspects of the environment, and also cases such as dispersal and migration (Odling-Smee, Laland, and Feldman 2003).
12. Although in later work he did allow organisms to compete in games and thereby develop more interesting behaviors (e.g., Reed, Toombs, and Barricelli 1967).
13. Examples of initial work in this area include McMullin and Varela 1997 and Hutton 2007.